\documentclass{article}

\PassOptionsToPackage{numbers}{natbib}  
\usepackage[final,dblblindworkshop]{neurips_2026}
\workshoptitle{GlobalSouthAI: Rethinking AI for and from the Global South}

\usepackage[utf8]{inputenc}
\usepackage[T1]{fontenc}
\usepackage{xcolor}    
\usepackage{url}
\usepackage{booktabs}
\usepackage{amsfonts}
\usepackage{amsmath,amssymb}
\usepackage{microtype}
\usepackage{graphicx}
\usepackage{float}
\usepackage{enumitem}
\usepackage{hyperref}

\title{In-Place Indic Tokenizer Expansion and Continual Pretraining of Nemotron-3-Nano-30B-A3B}

\author{%
  Avinash Benki\thanks{Equal contribution.}\quad
  Prasanjith Pasunuti\textsuperscript{*}\quad
  Sutanshu Saksena Raj\textsuperscript{*} \\
  \textbf{Parijat Chatterjee\quad Nayanjit Sarkar\quad Daksh Maan} \\
  Gnani.AI \\
  \texttt{\{avinash, prasanjith.p, sutanshu.raj\} @ gnani.ai}
}

\begin{document}

\maketitle

\begin{abstract}
We adapted Nemotron-3-Nano-30B-A3B for eight Indian languages without changing its backbone architecture.
We expanded the vocabulary in place from 131{,}072 to 163{,}072 entries using continued byte-pair encoding and AdaptMix rebalancing, reducing tokens per word by 38.6\%.
We initialized new input and output rows by averaging their original constituent rows, then trained them on 90\% Indic and 10\% English text before updating the backbone.
Continual pretraining then progressed through broad training, high-quality annealing, and long-context training over approximately 1.27 trillion tokens.
The annealed checkpoint gained 12.9 points on the Indic MILU benchmark but lost 3.4 on English MMLU.
We also report practical negative results. Direct token addition underperformed continued byte-pair encoding. AdamW weight decay changed rows intended to stay fixed. Perplexity on an Indic-heavy holdout did not select the learning \mbox{rate that best preserved English performance.}
\end{abstract}

\section{Introduction}

Tokenizers can split words in Indian languages into many small pieces, increasing the number of tokens needed to represent a document. We investigate whether expanding an existing tokenizer, followed by continual pretraining (CPT), can improve tokenization efficiency and \mbox{benchmark performance across eight Indian languages.}

We use Nemotron-3-Nano-30B-A3B Base BF16~\cite{nemotron3}, which has 31.75B parameters and activates 3.31B per token. We retain its hybrid backbone of Mamba-2~\cite{mamba2}, mixture-of-experts, and attention layers. Adding 32{,}000 vocabulary entries increases the untied input and output matrices by $2 \times 32{,}000 \times 2{,}688 = 172{,}032{,}000$ parameters. Table~\ref{tab:arch} gives the architecture details.

\paragraph{Relation to prior work.} We follow established methods for continued byte-pair encoding (BPE), initialization from existing token embeddings, embedding-only warmup, and subsequent CPT~\cite{inplace}. We use AdaptMix-style rebalancing to adjust each language's share of the tokenizer training data~\cite{adaptmix}. Our contribution is an account of applying these methods to eight Indian languages on a hybrid model, including the implementation problems and measured trade-offs.

\subsection{Contributions}
\begin{enumerate}[leftmargin=*,itemsep=2pt,topsep=2pt]
  \item We document tokenizer expansion and embedding warmup for eight Indian languages, including tokenization comparisons and checks that keep inherited vocabulary rows fixed (Sections~\ref{sec:tokenizer} and~\ref{sec:warmup}).
  \item We report continual pretraining over approximately 1.27T tokens across broad training, high-quality annealing, and long-context training, with evaluations of Indic gains, English retention, and long-context performance (Sections~\ref{sec:cpt} and~\ref{sec:cost}).
\end{enumerate}

\section{Tokenizer expansion}
\label{sec:tokenizer}

We first expanded the vocabulary from 131{,}072 to 163{,}072 entries using continued BPE~\cite{inplace}. This preserves existing token identities and adds merges for Hindi, Bengali, Gujarati, Kannada, Malayalam, Marathi, Tamil, and Telugu. The construction corpus also includes romanized text.

For a corpus $\mathcal{D}$, tokenizer fertility is $F(\mathcal{D}) = \sum_{x\in\mathcal{D}} N_{\mathrm{tok}}(x) / \sum_{x\in\mathcal{D}} N_{\mathrm{word}}(x)$, where $N_{\mathrm{tok}}(x)$ and $N_{\mathrm{word}}(x)$ count tokens and whitespace-delimited words in text $x$, respectively. Lower fertility means less fragmentation. Native-script comparisons use 2{,}000 Samanantar~\cite{samanantar} samples per language with seed 42. Fertility is computed separately for each of the eight target languages, and every mean we report across them is unweighted.

\paragraph{Expansion method.} In an initial comparison, continued BPE gave lower fertility than direct token addition with \texttt{add\_tokens} in all eight languages. Mean fertility was 2.17 for continued BPE and 2.48 for direct addition. We used continued BPE in all later experiments.

\paragraph{Normalization and rebalancing.} Normalization variants produced similar fertility scores. We selected NFKC with newline cleanup and Indic-aware preprocessing, although the boundary-matching check did not provide a controlled comparison of these variants. We then ran nine rounds of AdaptMix-style rebalancing~\cite{adaptmix}, increasing the data share of languages with higher fertility. The loop converged by round five. Appendix~\ref{app:tokenizer} gives the normalization results, quota updates, and final tokenizer preparation.

\paragraph{Final tokenizer.} The selected tokenizer, TOK-2, reduced mean fertility from 3.55 to 2.18 tokens per word, a 38.6\% reduction. Its score differs slightly from the initial continued-BPE result because it uses the selected normalization and rebalanced data. Gujarati improved most at 47.1\%, followed by Malayalam at 44.3\% (Table~\ref{tab:fertility}, Figure~\ref{fig:fertility}). These are tokenization results, not measured latency gains. The larger output vocabulary also changes per-token compute.

\paragraph{Romanized text.} Despite its inclusion during construction, romanized text remained more fragmented than native-script text. On paired transliterated samples, the mean romanized-to-native fertility ratio was 1.84. Tamil had the largest ratio at 2.58, with 5.96 tokens per romanized word compared with 2.31 per native-script word.

\section{Embedding warmup}
\label{sec:warmup}

We initialized and trained only the new input and output rows. Each new token $t$ was decomposed into a sequence of original subtokens $S(t)$. We used constituent-mean initialization,
\[
  e_{\mathrm{new}}(t) = \frac{1}{|S(t)|}\sum_{s\in S(t)} e_{\mathrm{old}}(s),
\]
where $e_{\mathrm{old}}(s)$ is the original row for subtoken $s$. We computed the mean separately for the input and output matrices. This follows source-subtoken averaging~\cite{inplace}, rather than the weighted-overlap method used by FOCUS~\cite{focus}. The final warmup used approximately 3B tokens, with 90\% Indic and 10\% English text, while keeping the backbone and inherited vocabulary rows fixed.

\paragraph{Fixing the frozen rows.} Inherited rows changed during an early warmup run despite gradient masking. We traced the drift to AdamW~\cite{adamw} weight decay, whose decay-only update is
\[
  \theta_{t+1} = \theta_t - \eta\lambda\theta_t,
\]
where $\eta$ is the learning rate and $\lambda$ is the weight-decay coefficient. This update changes inherited rows even when their gradients are masked to zero. Their relative drift reached approximately 4\% over 8{,}000 iterations. Training loss appeared normal, but English scores fell to near-random levels. We fixed this by saving the inherited rows before each optimizer step and restoring them afterwards. After each step, we verified \mbox{bit-exact equality on every tensor-parallel rank.}

\begin{figure}[t]
  \centering
  \includegraphics[width=0.78\linewidth]{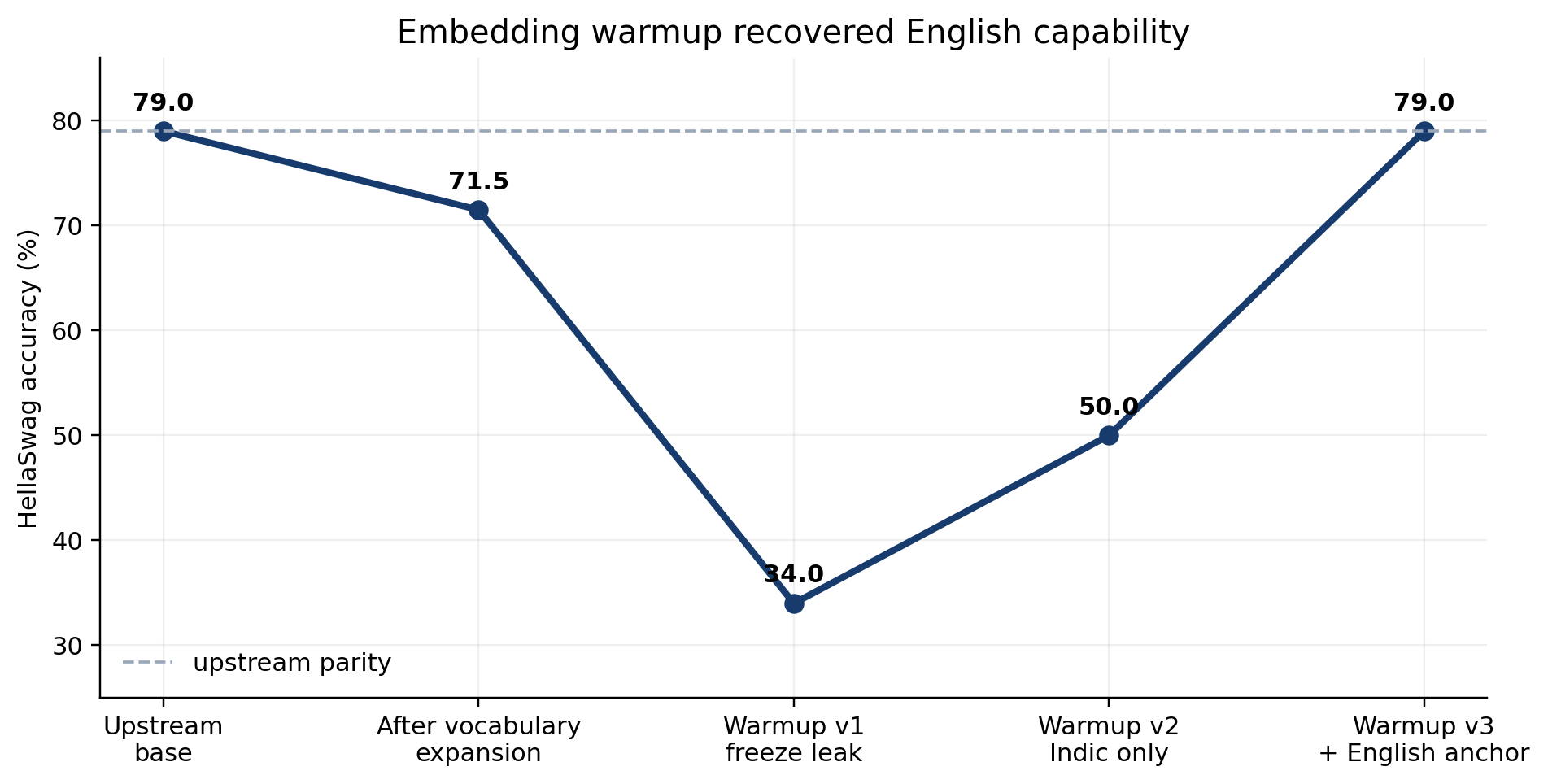}
  \caption{English scores during initialization and embedding warmup, shown in run order. These runs form a debugging sequence, not a controlled ablation. These scores are not directly comparable with the CPT results in Table~\ref{tab:cpteval}.}
  \label{fig:warmup}
\end{figure}

\paragraph{Including English data.} English scores recovered only partially after Indic-only warmup. A later run that included 10\% English text restored parity on the warmup evaluation (Figure~\ref{fig:warmup}). These runs were part of a sequential debugging process, not a controlled ablation. Without a matched comparison under the corrected freeze, we cannot isolate the effect of the English data.

\section{Continual pretraining}
\label{sec:cpt}

After embedding warmup, we continued pretraining with all model parameters trainable. Training had three stages: broad training at 8K context, high-quality annealing at 8K, and long-context training at 128K (Figure~\ref{fig:cptstages}). Together, they consumed approximately 1.27T tokens on 256 H200 GPUs.

\paragraph{Data preparation.} We combined English code, mathematics, reasoning, and high-quality web text with Indic web, academic, reference, and translated material from a pool of approximately 691B Indic tokens. Processing used exact and then inter-document fuzzy deduplication~\cite{dclm}. Indic documents received quality scores and high, medium, or low tiers based on length, script purity, repetition, boilerplate, KenLM perplexity~\cite{kenlm}, and safety signals. Synthetic translations and rewrites were down-weighted during broad training and excluded from high-quality annealing.

\paragraph{Learning-rate selection.} We compared four peak learning rates from $10^{-4}$ to $5{\times}10^{-6}$, training each run for 1B tokens. The lowest validation perplexity came from $10^{-4}$, while $10^{-5}$ best preserved English benchmark performance. We selected $10^{-5}$ with a Warmup-Stable-Decay schedule, a global batch size of 2{,}048, and a sequence length of 8{,}192 (Figure~\ref{fig:lr}). The validation set was Indic-heavy, so its aggregate perplexity did not capture the English retention trade-off.

\paragraph{Broad training and annealing.} Broad training consumed approximately 974.5B tokens at sequence length 8{,}192. During mixture planning, a separate 120B-token ablation with 55\% English and 45\% Indic data showed that gains varied across languages. Kannada gained 14.4 MILU points, Tamil 12.4, and Gujarati 12.2, while Hindi and English declined slightly. Increasing the Indic share to 60\% did not improve aggregate Indic performance. These results informed later allocation alongside source quality and repetition constraints (Appendix~\ref{app:ablation}, Figure~\ref{fig:milu}). High-quality annealing then continued from the selected broad checkpoint for 194.9B tokens at the same context length.

\paragraph{Long-context training.} The final stage used approximately 99.86B tokens of direct long-document training at sequence length 131{,}072. We evaluated this checkpoint on RULER~\cite{ruler} and question-answering tasks separately from the broad and annealed checkpoints. Model FLOP utilization was 24.95\% for broad training, 25.87\% for annealing, and 27.15\% at long context. Appendices~\ref{app:compute} and~\ref{app:cpttraces} record the compute costs and the training traces for all three stages.

\section{Results and English retention}
\label{sec:cost}

Table~\ref{tab:cpteval} compares all three checkpoints using the same evaluation setup.

\begin{table}[t]
\centering
\caption{Results under the same evaluation setup. Both adapted checkpoints precede long-context training. Appendix~\ref{app:fullcpt} lists the full benchmark results for these checkpoints.}
\label{tab:cpteval}
\small
\begin{tabular}{@{}lrrr@{}}
\toprule
\textbf{Benchmark} & \textbf{Base} & \textbf{Broad} & \textbf{Anneal} \\
\midrule
MILU~\cite{milu} & 40.63 & 53.54 & 53.52 \\
ARC-Challenge-Indic~\cite{arcindic} & 35.23 & 48.20 & 48.20 \\
IndicGLUE-selected~\cite{indicglue} & 52.12 & 57.86 & 57.22 \\
IndicMMLU-Pro~\cite{indicmmlupro} & 14.44 & 20.03 & 20.02 \\
BharatMath & 29.99 & 41.06 & 39.70 \\
Indic BPB overall ($\downarrow$)~\cite{indicgenbench} & 0.8453 & 0.6189 & 0.6132 \\
\midrule
MMLU~\cite{mmlu} & 78.50 & 75.20 & 75.11 \\
MMLU-Pro~\cite{mmlupro} & 64.83 & 54.78 & 55.43 \\
HellaSwag~\cite{hellaswag} & 85.57 & 83.52 & 83.69 \\
\bottomrule
\end{tabular}
\end{table}

Relative to the base model, the annealed checkpoint gained 12.9 points on MILU and 13.0 on ARC-Challenge-Indic. Indic bits per byte fell from 0.8453 to 0.6132. MMLU declined by 3.4 points and MMLU-Pro by 9.4. Annealing recovered part of the English cost of broad training: the nine-benchmark English mean rose from 67.98 to 68.40, with MMLU-Pro regaining 0.7 points while MMLU held flat.

We selected the annealed checkpoint for downstream post-training because it maintained the main Indic gains, further reduced Indic bits per byte, and recovered part of the English retention cost. A different balance between Indic and English performance could still favor the broad checkpoint. The warmup scores in Figure~\ref{fig:warmup} use a different setup and cannot \mbox{be compared directly with the CPT evaluations.}

\paragraph{Long-context evaluation.} The long-context checkpoint achieved mean RULER recall of 76.7\% across 55 evaluation slices. Mean exact match was 45.7\% across 10 question-answering slices. Recall decreased from 96.6\% at 8{,}192 tokens to 46.9\% at 131{,}072 tokens. Appendix~\ref{app:longeval} gives recall at each of the five evaluated context lengths.

A companion HELMET~\cite{helmet} evaluation shows that long-context performance varies by task. In-context learning improves from 59.2 at 8K to 81.7 at 128K and is the only evaluated category whose 128K score exceeds its 8K score. Retrieval-augmented generation and long-form question answering decline to 37.2 and 26.8 at 128K, respectively; long-form question answering peaks at 40.6 at 64K. Exact retrieval deteriorates markedly beyond 32K, falling from 95.9 at 8K to 14.0 at 128K.

\section{Limitations and conclusion}

We did not run a matched CPT control with the original tokenizer, so we cannot separate the contribution of tokenizer expansion from additional training. The warmup sequence was not a controlled ablation. The pretraining corpus includes twelve Indic languages, but only eight received dedicated vocabulary expansion. Odia, Punjabi, Urdu, and Nepali received a small share of the corpus without expansion. Romanized text remained more fragmented, and we did not measure end-to-end latency. The long-context results report one checkpoint without a matched base-model comparison.

Tokenizer expansion and CPT improved native-script tokenization efficiency and Indic benchmark performance, but reduced accuracy on several English benchmarks. Our experience also shows why embedding warmup needs explicit checks that inherited rows remain unchanged, and why model selection should measure English retention alongside Indic performance.

\paragraph{Model release.} The downstream model is available at \url{https://huggingface.co/gnani/gnani-evon-v3.3-30B-A3B}. It includes supervised fine-tuning and reinforcement learning beyond the CPT stages evaluated here.

\clearpage
\bibliographystyle{plain}
\begingroup
\raggedright

\endgroup

\clearpage
\appendix
\begingroup
\raggedbottom
\widowpenalty=10000
\clubpenalty=10000

\section{Tokenizer experiments}
\label{app:tokenizer}

Table~\ref{tab:tokenizersweep} summarizes the tokenizer comparisons recorded in the report compiled on July 13, 2026. We first compared expansion methods, then normalization and language rebalancing. TOK-2 was selected for training. The larger TOK-3 tokenizer was tested later and was not used for CPT.

\begin{table}[H]
\centering
\caption{Tokenizer fertility on Samanantar, using seed 42 and 2{,}000 samples per language.}
\label{tab:tokenizersweep}
\small
\begin{tabular}{@{}llrr@{}}
\toprule
\textbf{Run} & \textbf{Extension} & \textbf{Mean fertility} & \textbf{Notes} \\
\midrule
Base Nemotron & None & 3.5472 & Original tokenizer \\
Direct add\_tokens & 32K & 2.4815 & 14.3\% above continued BPE \\
Continued BPE & 32K & 2.1715 & Used in subsequent experiments \\
Raw / NFKC / NFKC+regex & 32K & 2.1716--2.1823 & Similar fertility across variants \\
TOK-2 & 32K & 2.1773 & Selected for CPT \\
TOK-3 & 64K & 2.0996 & Lowest mean, not used for CPT \\
\bottomrule
\end{tabular}
\end{table}

\paragraph{Normalization comparison.}
We compared raw text, Unicode NFKC, and NFKC followed by newline cleanup and an Indic-aware regex pass on the same 7.58M-row mixture. Mean fertility was 2.1716 for raw text and 2.1823 for both normalized variants. Fertility did not distinguish the two normalized variants, and we selected the regex variant. A boundary-agreement check against the base pretokenizer was run only for the selected variant. Variants without a regex stage match the base pretokenizer by construction, so this check cannot rank the three variants. We therefore have no measured criterion that distinguishes them.

\paragraph{Rebalancing rule.}
We rebalanced language shares using these fertility-based updates:
\[
\delta_l = \frac{f_l - f_{\mathrm{best}}}{f_{\mathrm{range}}},
\qquad w_l = \max(\delta_l + \epsilon, \epsilon),
\qquad m_l = (1-\mu)m_{l,\mathrm{prev}} + \mu t_l.
\]
Here $f_l$ is the fertility of language $l$, $f_{\mathrm{best}}$ is the lowest fertility, and $f_{\mathrm{range}}$ is the range across languages. The update moves the previous data share $m_{l,\mathrm{prev}}$ toward the target share $t_l$. We used $\mu=0.4$ and $\epsilon=0.001$ in all nine rebalancing rounds.

\paragraph{Changes in language shares.}
From initialization to round nine, Malayalam's share rose from 12.5\% to 18.7\%, Kannada's from 12.5\% to 14.3\%, and Telugu's from 12.5\% to 14.5\%. Hindi's share fell from 12.5\% to 7.0\% as more data was allocated to languages with higher fertility.

\paragraph{Rebalancing result.}
The nine-round loop converged by round five. Malayalam had the highest fertility before rebalancing. Its data share increased from 12.5\% to 18.7\%, and its fertility improved by 5.4\% relative to the initial tokenizer.

\paragraph{Final tokenizer preparation.}
Of the 32{,}000 appended entries, 31{,}999 were byte-identical between the rebalanced artifact and the tokenizer staged for training. A space-plus-newline merge was made unreachable and its entry was later reused as an atomic separator token. This kept the \mbox{vocabulary size and tensor alignment unchanged.}

\paragraph{Later 64K expansion.}
The exploratory TOK-3 tokenizer reduced fertility relative to the original tokenizer by 27.2\% for Hindi, 51.8\% for Gujarati, 43.5\% for Kannada, and 42.5\% for Malayalam. Compared with TOK-2, its mean fertility was 3.6\% lower, but Malayalam fertility was 3.1\% higher (Figure~\ref{fig:mixv5}). The CPT configuration was fixed before TOK-3 was developed, and all runs used TOK-2.

\clearpage
\section{CPT configuration}
\label{app:config}

\begin{table}[H]
\centering
\caption{Configuration used for broad continual pretraining.}
\small
\begin{tabular}{@{}ll@{}}
\toprule
\textbf{Field} & \textbf{Value} \\
\midrule
Initialization & Checkpoint after warmup, vocabulary 163{,}072 \\
Precision & BF16 \\
Sequence length & 8{,}192 (broad and anneal), 131{,}072 (long context) \\
Global / micro batch & 2{,}048 / 1 \\
Parallelism & TP=1, EP=8, PP=1, remaining ranks use DP \\
Optimizer & Distributed AdamW \\
Peak / min learning rate & $1.0{\times}10^{-5}$ / $1.0{\times}10^{-6}$ \\
Schedule & Warmup-Stable-Decay with square-root decay \\
Warmup iterations & 200 \\
MoE dispatcher & flex + HybridEP, capacity factor 1.25 \\
Kernels & Grouped GEMM, selective recompute, CUDA graphs \\
Tracking & Weights \& Biases \\
\bottomrule
\end{tabular}
\end{table}

\section{Launch configuration and validation}
\label{app:gpu16}

The CPT setup uses Nemotron-Bridge/Megatron, with training scripts, data-mixture manifests, and pipelines for building the caches used during model training.

\paragraph{Full training setup.}
Training used 32 nodes with eight H200 GPUs per node, for 256 GPUs in total. The parallelism setting was \texttt{REPLICA\_MP = (1, 8, 1)}, with expert-parallel groups kept within each node. Dense data parallelism was 256 and expert data parallelism was 32. We used eight gradient-accumulation steps with a microbatch size of one.

\paragraph{Mid-training data plan.}
The mid-training manifest specifies 285B tokens over 16{,}987 iterations, with global batch size 2{,}048 and sequence length 8{,}192. Approximately 34\% of tokens come from long-document sources. The remaining tokens cover shorter examples in reasoning, code, knowledge retention, and instruction data. Table~\ref{tab:285b} lists the planned rather than measured token allocation.

\paragraph{Validation on 128 GPUs.}
We tested the launch setup on 16 nodes using the checkpoint after embedding warmup and the tokenizer-construction data mixture described above. The run completed 100 iterations with a peak learning rate of $10^{-5}$ and a Warmup-Stable-Decay stable fraction of 0.80. During iterations 50--100, step time was 11.4--11.6\,s and compute throughput was 254--261 TFLOP/s/GPU. Language-model loss fell from 4.78 at iteration 10 to 4.05 during iterations 50--100. Gradient norm was approximately 0.9--1.6. The run had no skipped or NaN iterations. We saved the \mbox{checkpoint after completing all 100 iterations.}

\paragraph{Full-run settings.}
The 32-node configuration sets the vocabulary size to 163{,}072 and disables forced router load balancing. It also uses NCCL settings tuned for the InfiniBand interconnect.

\section{English and Indic data-mixture ablation}
\label{app:ablation}

\begin{table}[H]
\centering
\caption{MILU scores before and after 120B tokens of CPT with 55\% English and 45\% Indic data.}
\small
\begin{tabular}{@{}lrrr@{}}
\toprule
\textbf{Language} & \textbf{Base} & \textbf{120B CPT} & \textbf{$\Delta$} \\
\midrule
Kannada & 32.8 & 47.2 & $+$14.4 \\
Tamil & 34.2 & 46.6 & $+$12.4 \\
Gujarati & 34.4 & 46.5 & $+$12.2 \\
Telugu & 35.8 & 47.9 & $+$12.1 \\
Malayalam & 35.5 & 45.7 & $+$10.1 \\
Bengali & 39.4 & 47.7 & $+$8.4 \\
Marathi & 40.1 & 46.8 & $+$6.7 \\
Hindi & 53.1 & 51.4 & $-$1.8 \\
English & 65.6 & 63.9 & $-$1.7 \\
\bottomrule
\end{tabular}
\end{table}

The results informed later language allocation. The planned shares were 16\% Telugu, 15\% Tamil, 13\% Kannada, 12\% each for Gujarati, Malayalam, Bengali, and Hindi, and 8\% Marathi. These are planning weights, not measured shares of the completed training run.

\section{Long-context CPT mixture}
\label{app:longcontext}

\begin{table}[H]
\centering
\caption{Composition of the 100B-token long-context data pool. The selected run consumed approximately 99.86B tokens from this pool.}
\small
\begin{tabular}{@{}lrrr@{}}
\toprule
\textbf{Bucket} & \textbf{Weight} & \textbf{Shards} & \textbf{Pool tokens} \\
\midrule
Indic long documents & 0.22 & 9 & 22.0B \\
Indic high-quality & 0.15 & 132 & 15.0B \\
English long documents & 0.20 & 7 & 20.0B \\
Mathematics and STEM & 0.14 & 1 & 14.0B \\
Long-context supervised data & 0.10 & 3 & 10.0B \\
Code & 0.08 & 1 & 8.0B \\
Base-retention data & 0.07 & 2 & 7.0B \\
Other multilingual & 0.04 & 2 & 4.0B \\
\bottomrule
\end{tabular}
\end{table}

\section{Full checkpoint evaluation results}
\label{app:fullcpt}

\begin{table}[H]
\centering
\caption{Results for the base, broad, and annealed checkpoints. Two benchmarks are omitted because they lack results for one or more checkpoints.}
\small
\begin{tabular}{@{}lrrr@{}}
\toprule
\textbf{Benchmark} & \textbf{Base} & \textbf{Broad} & \textbf{Anneal} \\
\midrule
MILU~\cite{milu} & 40.63 & 53.54 & 53.52 \\
ARC-Challenge-Indic~\cite{arcindic} & 35.23 & 48.20 & 48.20 \\
Belebele~\cite{belebele} & 37.49 & 42.17 & 42.27 \\
BharatMath & 29.99 & 41.06 & 39.70 \\
Global-MMLU-Lite~\cite{globalmmlu} & 32.50 & 36.17 & 36.00 \\
IndicGLUE-selected~\cite{indicglue} & 52.12 & 57.86 & 57.22 \\
Indic BPB overall ($\downarrow$)~\cite{indicgenbench} & 0.8453 & 0.6189 & 0.6132 \\
IndicMMLU-Pro~\cite{indicmmlupro} & 14.44 & 20.03 & 20.02 \\
IndicParam~\cite{indicparam} & 28.08 & 31.59 & 31.55 \\
Sanskriti~\cite{sanskriti} & 84.15 & 80.60 & 80.82 \\
ARC-Challenge~\cite{arc} & 67.66 & 67.49 & 67.92 \\
GSM8K~\cite{gsm8k} & 86.13 & 84.46 & 84.46 \\
HellaSwag~\cite{hellaswag} & 85.57 & 83.52 & 83.69 \\
MMLU~\cite{mmlu} & 78.50 & 75.20 & 75.11 \\
MMLU-Pro~\cite{mmlupro} & 64.83 & 54.78 & 55.43 \\
OpenBookQA~\cite{openbookqa} & 46.60 & 43.60 & 44.80 \\
PIQA~\cite{piqa} & 84.44 & 83.03 & 83.46 \\
RACE-filtered~\cite{race} & 42.39 & 41.72 & 42.39 \\
WinoGrande~\cite{winogrande} & 79.48 & 78.06 & 78.37 \\
\bottomrule
\end{tabular}
\end{table}

The broad checkpoint generated an unintended code separator during HumanEval~\cite{humaneval}. Its HumanEval result therefore does not provide a reliable measure of code-generation retention.

\section{Long-context checkpoint evaluation}
\label{app:longeval}

We evaluated the long-context checkpoint on RULER~\cite{ruler} and question-answering (QA) tasks. A companion HELMET~\cite{helmet} evaluation measured additional long-context capabilities. Table~\ref{tab:longeval} reports the mean recall over 55 RULER slices, the mean exact match over 10 question-answering slices, and recall at five context lengths. The percentages are scores for this checkpoint rather than improvements over a reference model.

\begin{table}[H]
\centering
\caption{RULER recall and QA exact-match scores for the long-context checkpoint. These tasks differ from those used for the broad and annealed checkpoint comparisons.}
\label{tab:longeval}
\small
\begin{tabular}{@{}lr@{}}
\toprule
\textbf{Metric or context length} & \textbf{Score (\%)} \\
\midrule
Mean RULER recall (55 slices) & 76.7 \\
Mean QA exact match (10 slices) & 45.7 \\
\midrule
Recall at 8{,}192 tokens & 96.6 \\
Recall at 16{,}384 tokens & 90.6 \\
Recall at 32{,}768 tokens & 82.4 \\
Recall at 65{,}536 tokens & 67.3 \\
Recall at 131{,}072 tokens & 46.9 \\
\bottomrule
\end{tabular}
\end{table}

HELMET's Recall category scores are 95.9, 81.4, 57.0, 37.3, and 14.0 at the same five context lengths. The gap relative to RULER grows from 0.7 points at 8K to 32.9 points at 128K. The evaluations use different task subsets and aggregation schemes, so their recall scores should not be compared directly.

\section{CPT compute usage}
\label{app:compute}

\begin{table}[H]
\centering
\caption{GPU-hours for CPT, excluding embedding warmup and ablations.}
\small
\begin{tabular}{@{}lr@{}}
\toprule
\textbf{Stage} & \textbf{GPU-hours} \\
\midrule
CPT broad (${\sim}$974B) & ${\sim}$25{,}147 \\
CPT annealing (195B) & ${\sim}$4{,}988 \\
CPT long context (${\sim}$100B) & 4{,}977 \\
\midrule
\textbf{CPT total} & ${\sim}$35{,}112 \\
\bottomrule
\end{tabular}
\end{table}

Broad training and annealing ran at a median 5.85\,s per step on 256 GPUs, processing approximately 2.87M tokens per second. At sequence length 131{,}072, the median step time was 159.89\,s.

\clearpage
\section{Supporting tables and figures}
\label{app:floats}

This section provides the architecture summary, per-language fertility results, planned mid-training mixture, and additional plots for the tokenizer and CPT experiments.

\begin{table}[H]
\centering
\caption{Nemotron-3-Nano-30B-A3B architecture and expanded vocabulary. Parameter counts refer to the base model. Vocabulary expansion adds approximately 172 million parameters.}
\label{tab:arch}
\small
\begin{tabular}{@{}ll@{}}
\toprule
\textbf{Field} & \textbf{Value} \\
\midrule
Total / active parameters & 31.75B / 3.31B \\
Layers & 52 (23 Mamba-2, 23 MoE, 6 GQA) \\
Model dim / heads / KV heads & 2{,}688 / 32 / 2 \\
MoE experts (routed / active / shared) & 128 / 6 / 1 \\
Vocabulary / max sequence & 163{,}072 / 131{,}072 \\
Position encoding & none (NoPE) \\
\bottomrule
\end{tabular}
\end{table}

\begin{table}[H]
\centering
\caption{Fertility of the original tokenizer and TOK-2 (163{,}072 entries).}
\label{tab:fertility}
\small
\begin{tabular}{@{}lrrr@{}}
\toprule
\textbf{Language} & \textbf{Upstream} & \textbf{Expanded} & \textbf{Change} \\
\midrule
Hindi & 2.05 & 1.66 & $-$19.2\% \\
Bengali & 3.06 & 2.14 & $-$30.0\% \\
Gujarati & 3.61 & 1.91 & $-$47.1\% \\
Kannada & 3.86 & 2.34 & $-$39.5\% \\
Malayalam & 4.97 & 2.77 & $-$44.3\% \\
Marathi & 3.10 & 1.94 & $-$37.4\% \\
Tamil & 3.77 & 2.29 & $-$39.1\% \\
Telugu & 3.96 & 2.37 & $-$40.1\% \\
\midrule
Mean & 3.55 & 2.18 & $-$38.6\% \\
\bottomrule
\end{tabular}
\end{table}

\begin{table}[H]
\centering
\caption{Planned data allocation for the 285B mid-training mixture.}
\label{tab:285b}
\small
\begin{tabular}{@{}lrr@{}}
\toprule
\textbf{Bucket} & \textbf{Weight} & \textbf{Target tokens} \\
\midrule
Reasoning & 22\% & 62.7B \\
Indic long documents & 15\% & 42.8B \\
Math \& STEM & 14\% & 39.9B \\
Knowledge retention & 12\% & 34.2B \\
English long documents & 11\% & 31.4B \\
Code & 9\% & 25.7B \\
Instruction & 7\% & 20.0B \\
Indic high-quality & 6\% & 17.1B \\
Web / multilingual & 4\% & 11.4B \\
\bottomrule
\end{tabular}
\end{table}

\begin{figure}[H]
  \centering
  \includegraphics[width=\linewidth]{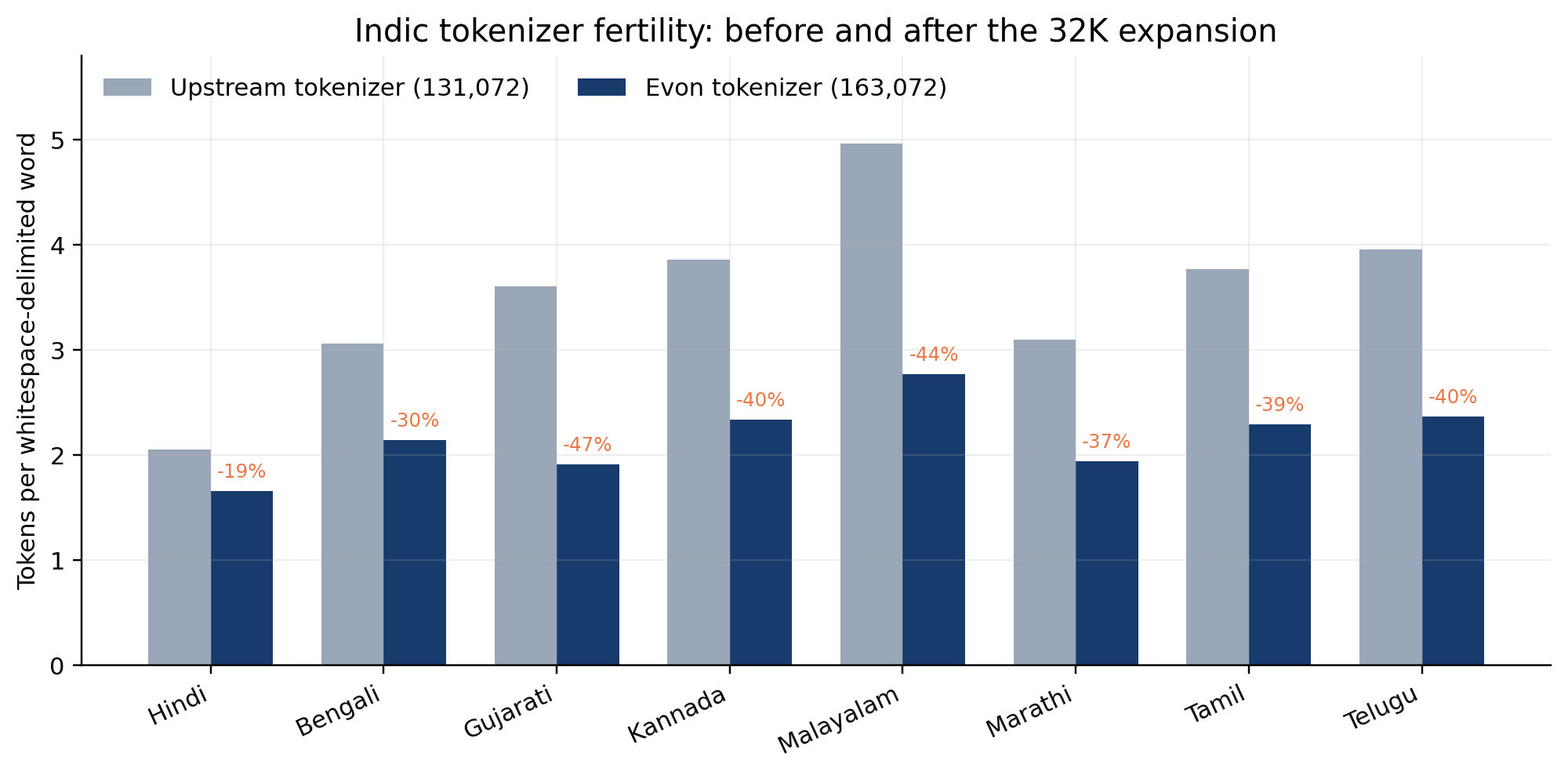}
  \caption{Tokens per word before and after the 32K expansion. Lower is better.}
  \label{fig:fertility}
\end{figure}

\begin{figure}[H]
  \centering
  \includegraphics[width=\linewidth]{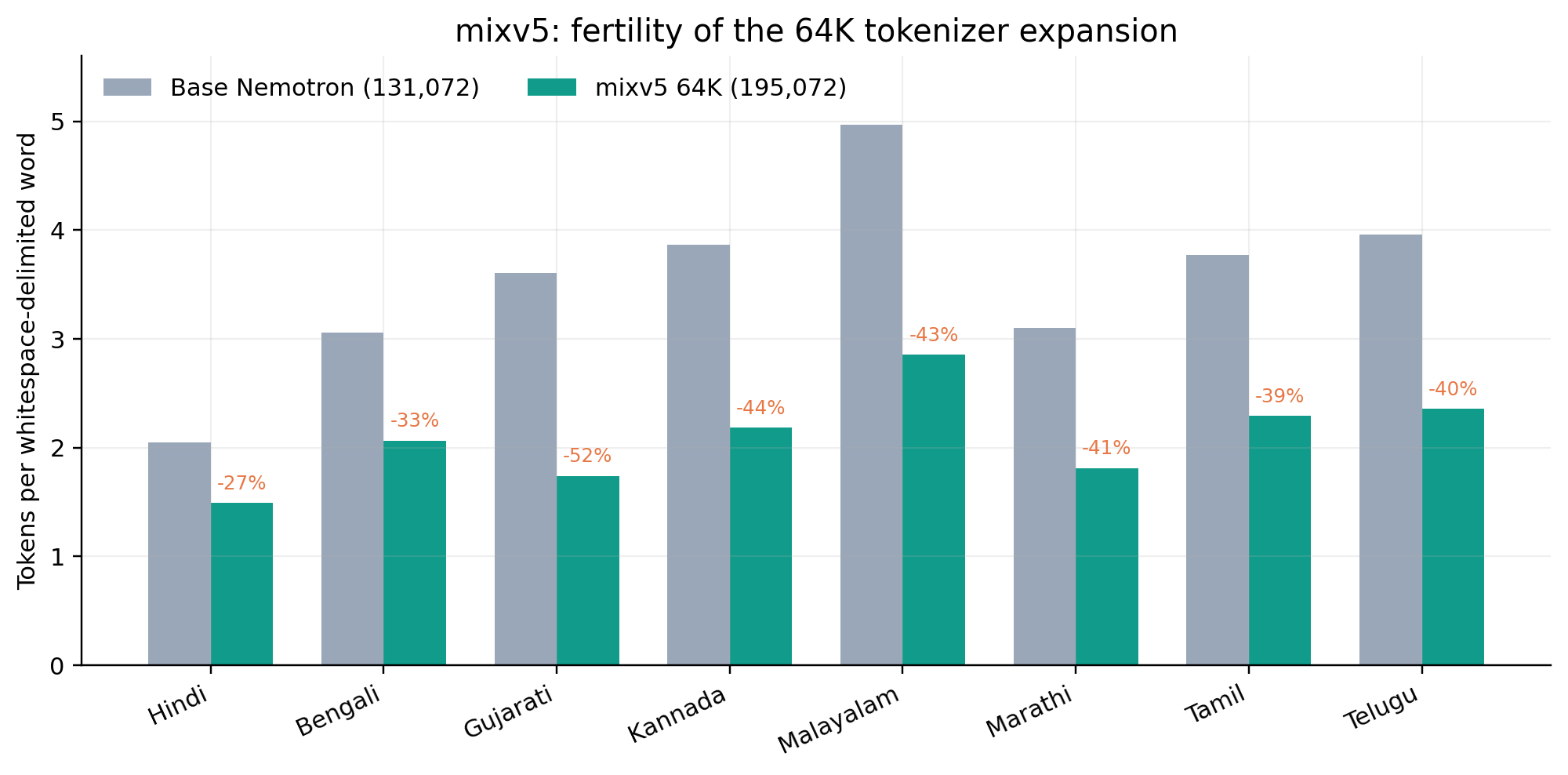}
  \caption{Fertility of the original tokenizer and the exploratory TOK-3 tokenizer with 64{,}000 added entries. TOK-3 was not used in any CPT run.}
  \label{fig:mixv5}
\end{figure}

\begin{figure}[H]
  \centering
  \includegraphics[width=0.88\linewidth]{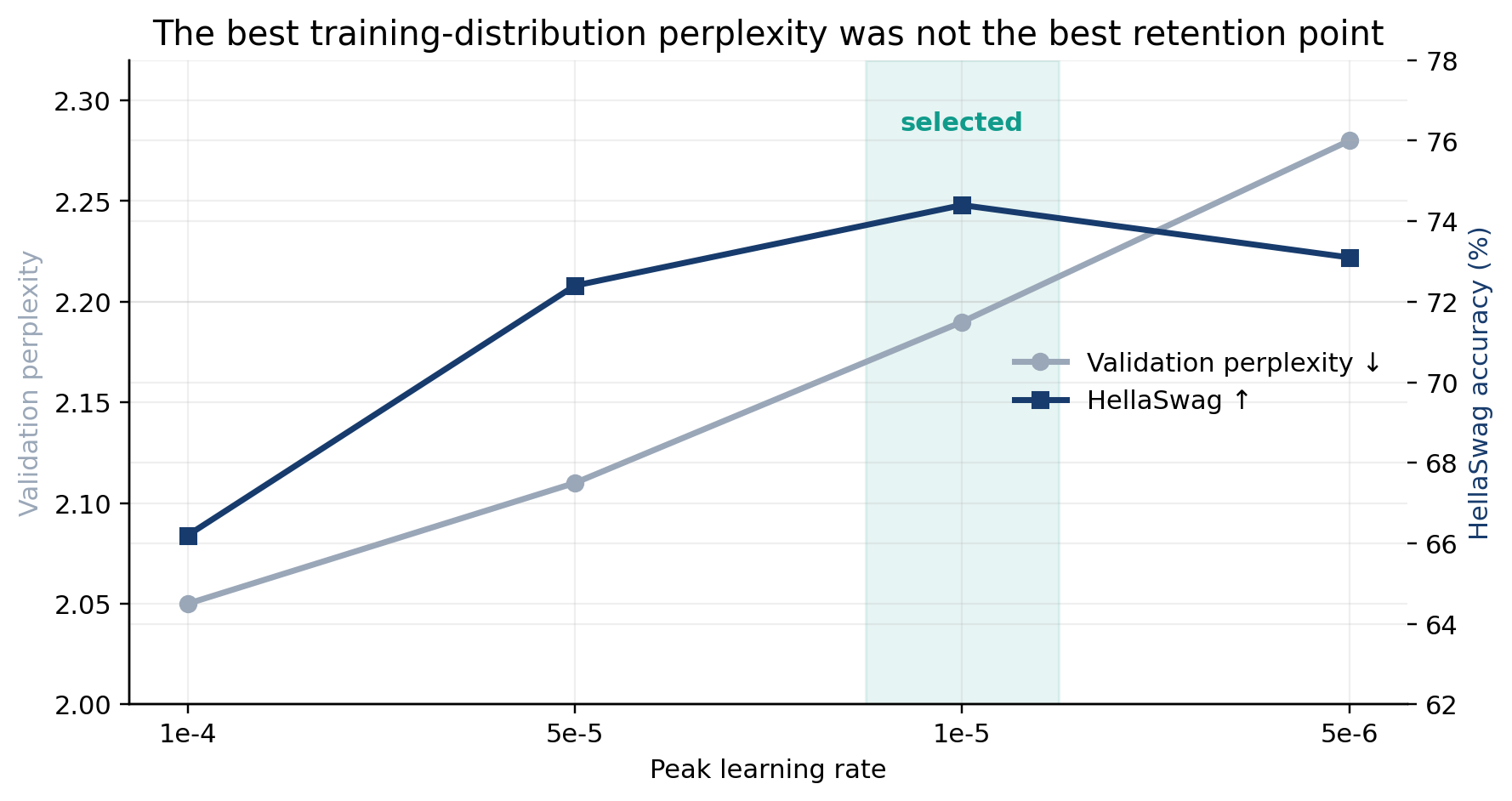}
  \caption{Learning-rate comparison. Validation perplexity and English benchmark performance favored different learning rates in the four-run sweep.}
  \label{fig:lr}
\end{figure}

\begin{figure}[H]
  \centering
  \includegraphics[width=0.72\linewidth]{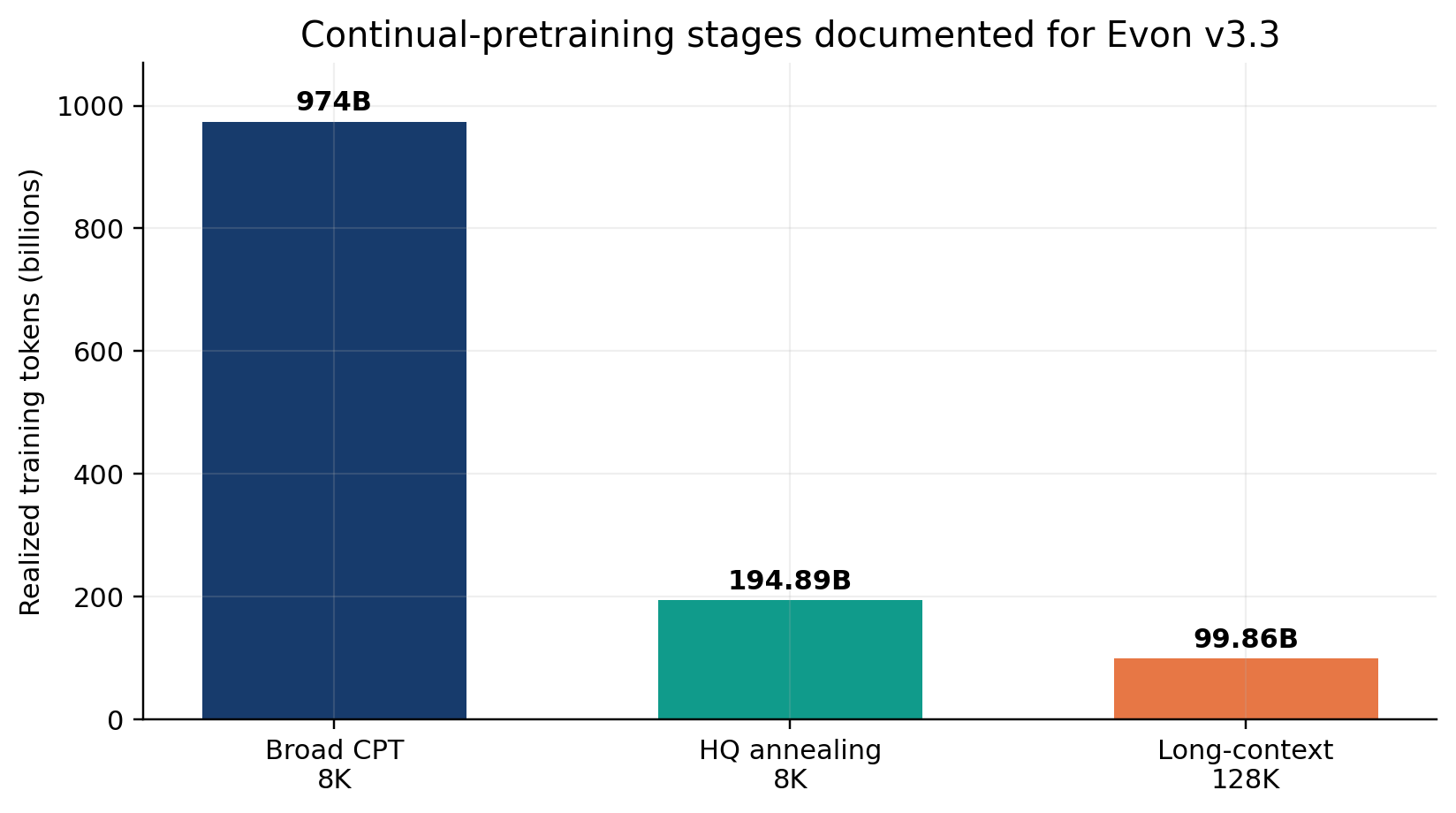}
  \caption{Tokens consumed during broad training, high-quality annealing, and long-context training.}
  \label{fig:cptstages}
\end{figure}

\begin{figure}[H]
  \centering
  \includegraphics[width=0.94\linewidth]{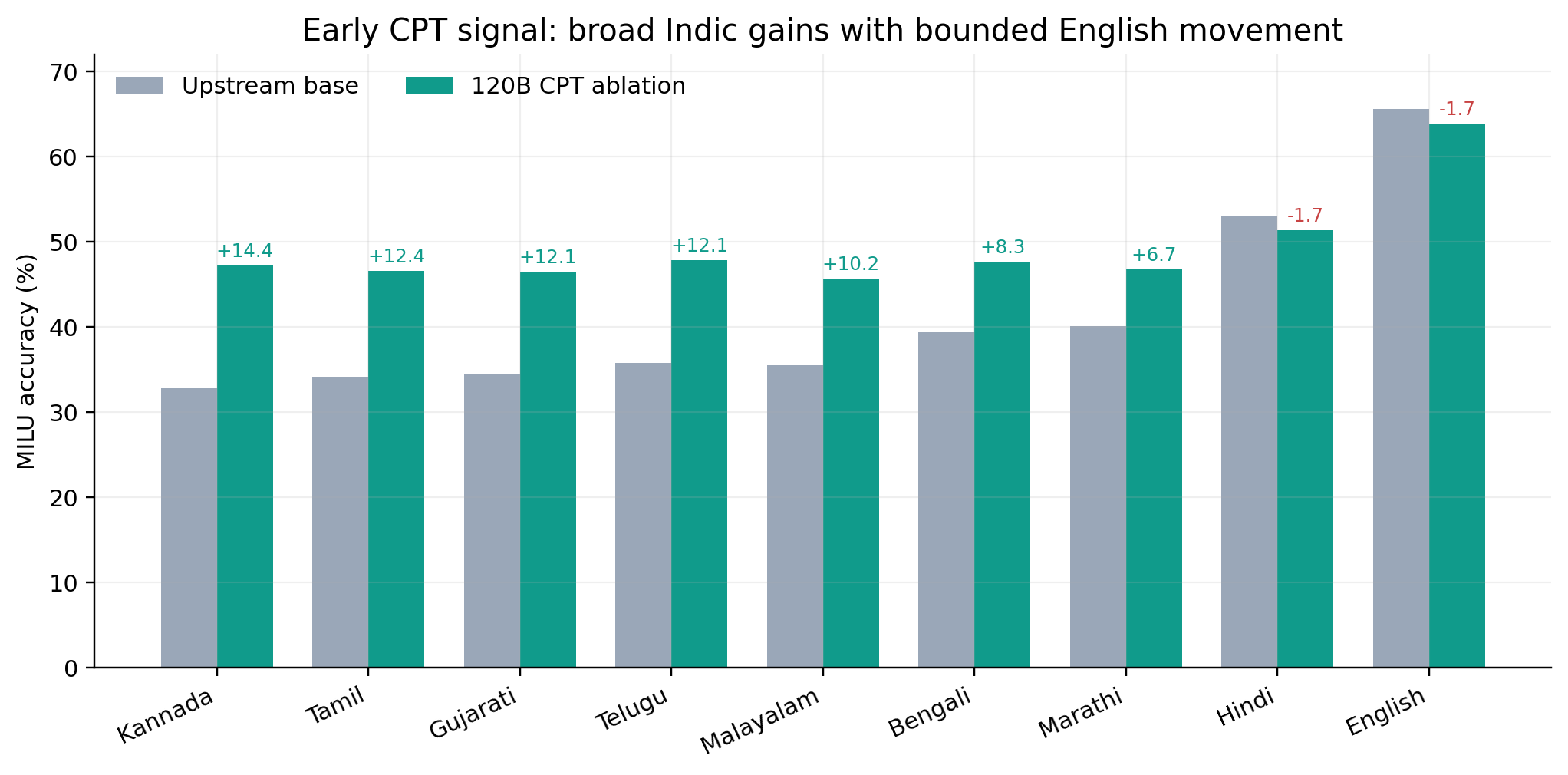}
  \caption{MILU accuracy before and after the 120B-token ablation with 55\% English and 45\% Indic data. This run informed mixture planning and was not released as a checkpoint.}
  \label{fig:milu}
\end{figure}

\begin{figure}[H]
  \centering
  \includegraphics[width=0.98\linewidth]{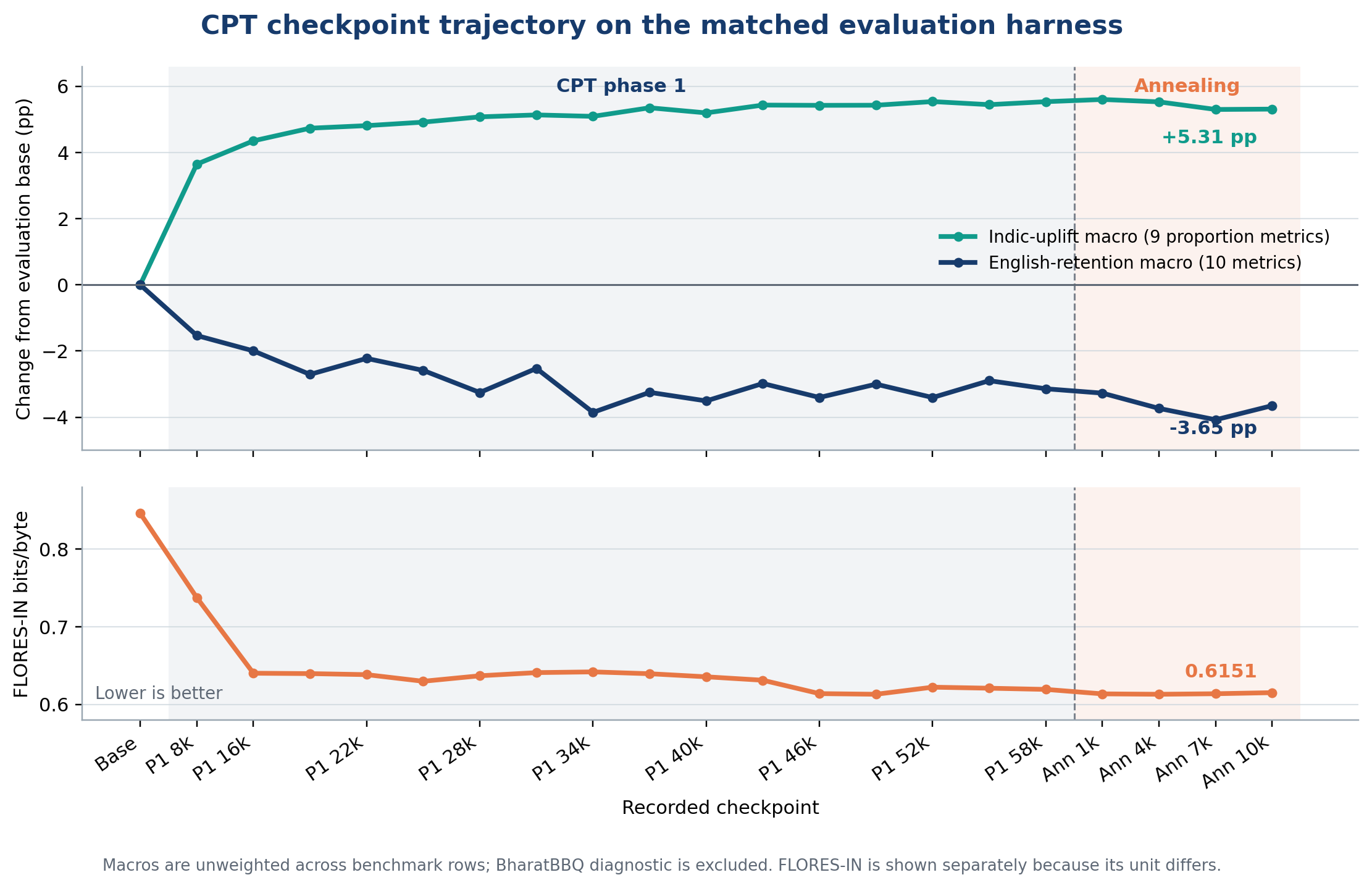}
  \caption{Indic and English benchmark results across CPT checkpoints under the same evaluation setup. Lower FLORES-IN bits per byte indicate better performance on that metric.}
  \label{fig:cpttraj}
\end{figure}

\clearpage
\section{CPT training traces}
\label{app:cpttraces}

The following plots were exported from the recorded CPT runs and are reproduced from the technical report. Each row shows MoE sequence load-balancing loss, gradient norm, learning rate, and validation language-model loss, from left to right. The load-balancing loss is an auxiliary objective and is separate from the language-model loss.

\begin{figure}[H]
  \centering
  \includegraphics[width=0.98\textwidth]{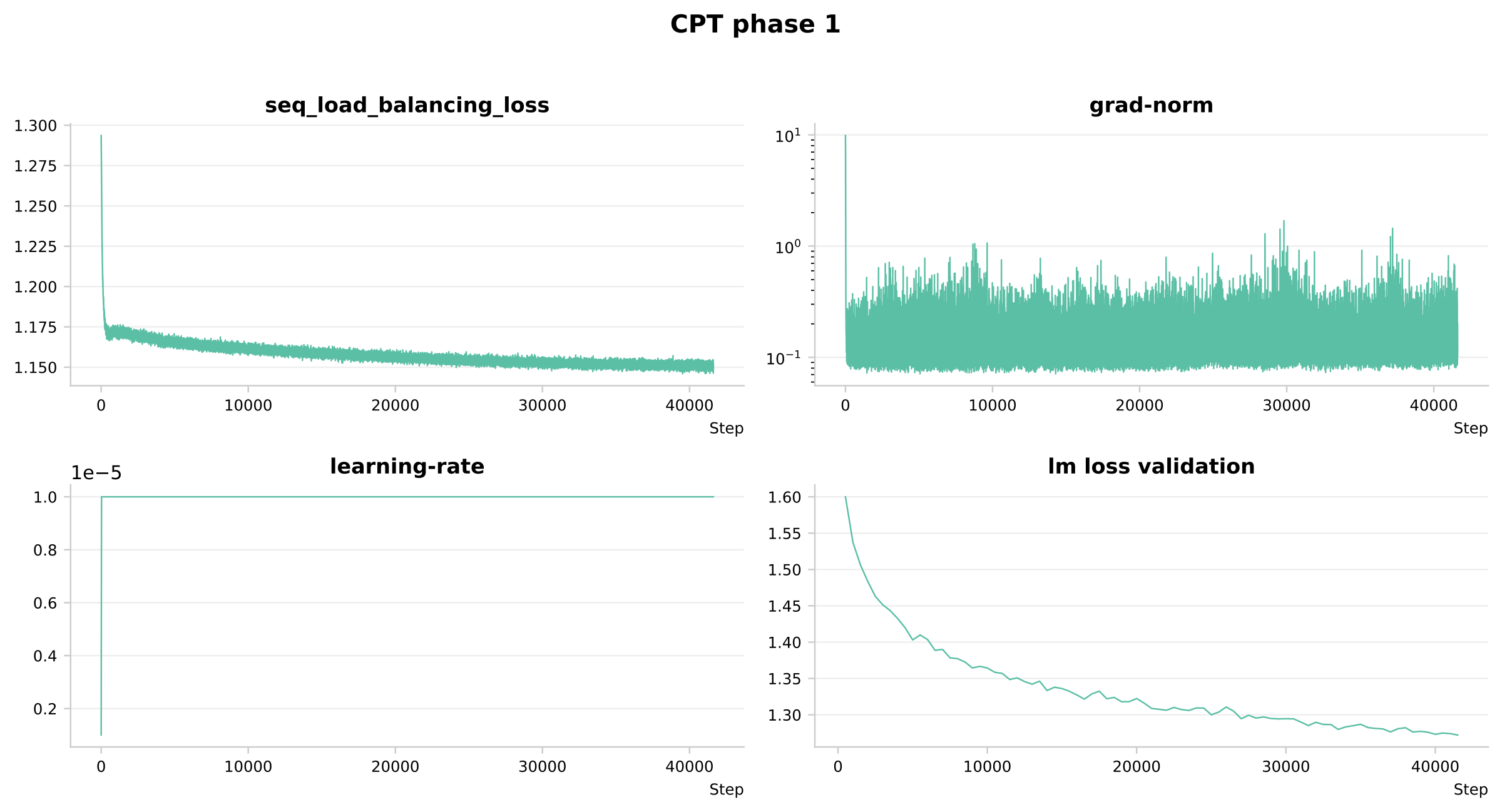}
  \caption{Weights \& Biases traces for broad training at 8K context. These plots end before the broad checkpoint reported in the benchmark evaluation.}
  \label{fig:cptphase1wandb}
\end{figure}

\begin{figure}[H]
  \centering
  \includegraphics[width=0.98\textwidth]{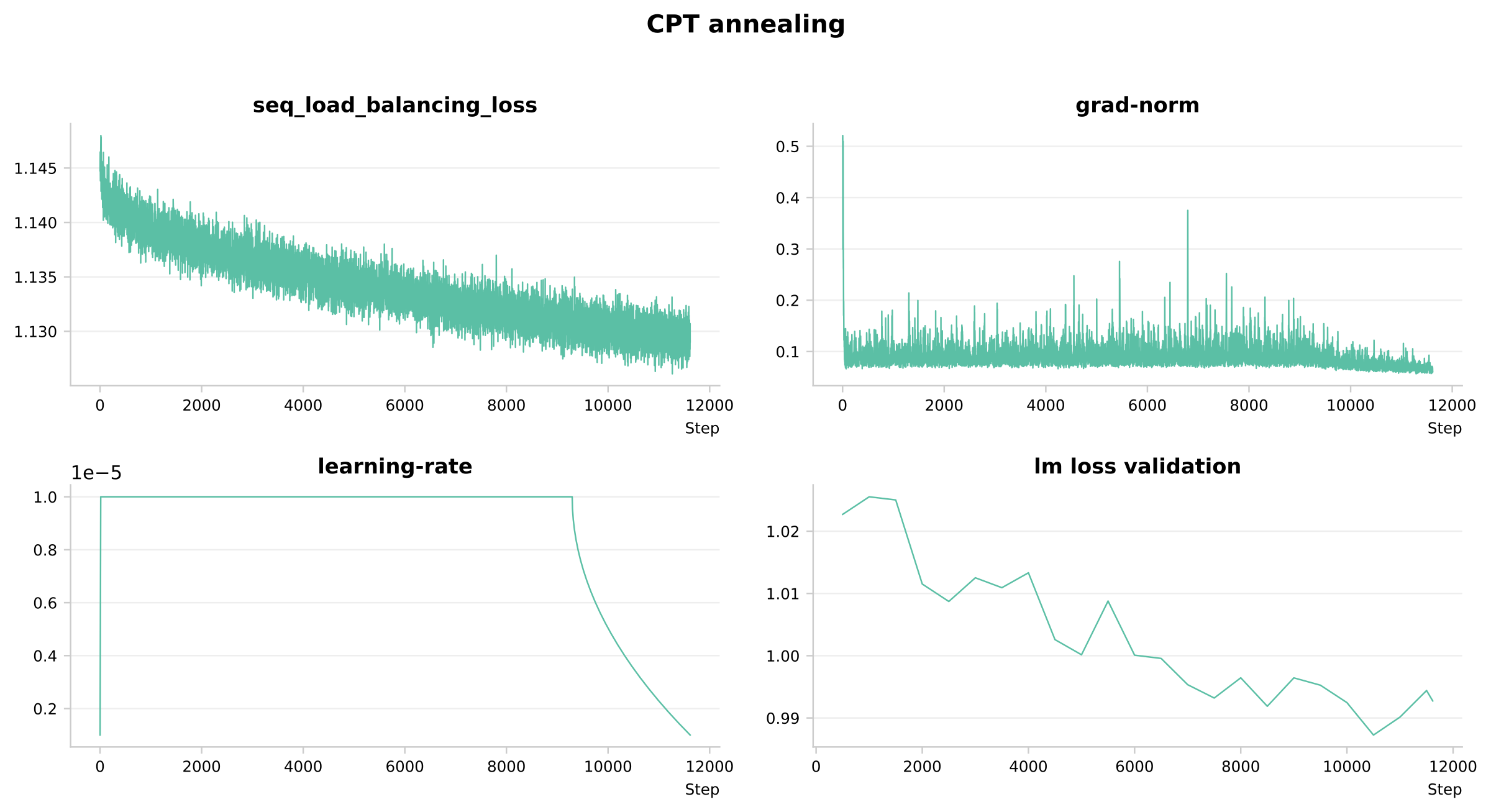}
  \caption{Weights \& Biases traces for high-quality annealing. Learning-rate decay begins late in this stage. Validation loss fluctuates around 1.0 over the recorded interval.}
  \label{fig:cptannealwandb}
\end{figure}

\begin{figure}[H]
  \centering
  \includegraphics[width=0.98\textwidth]{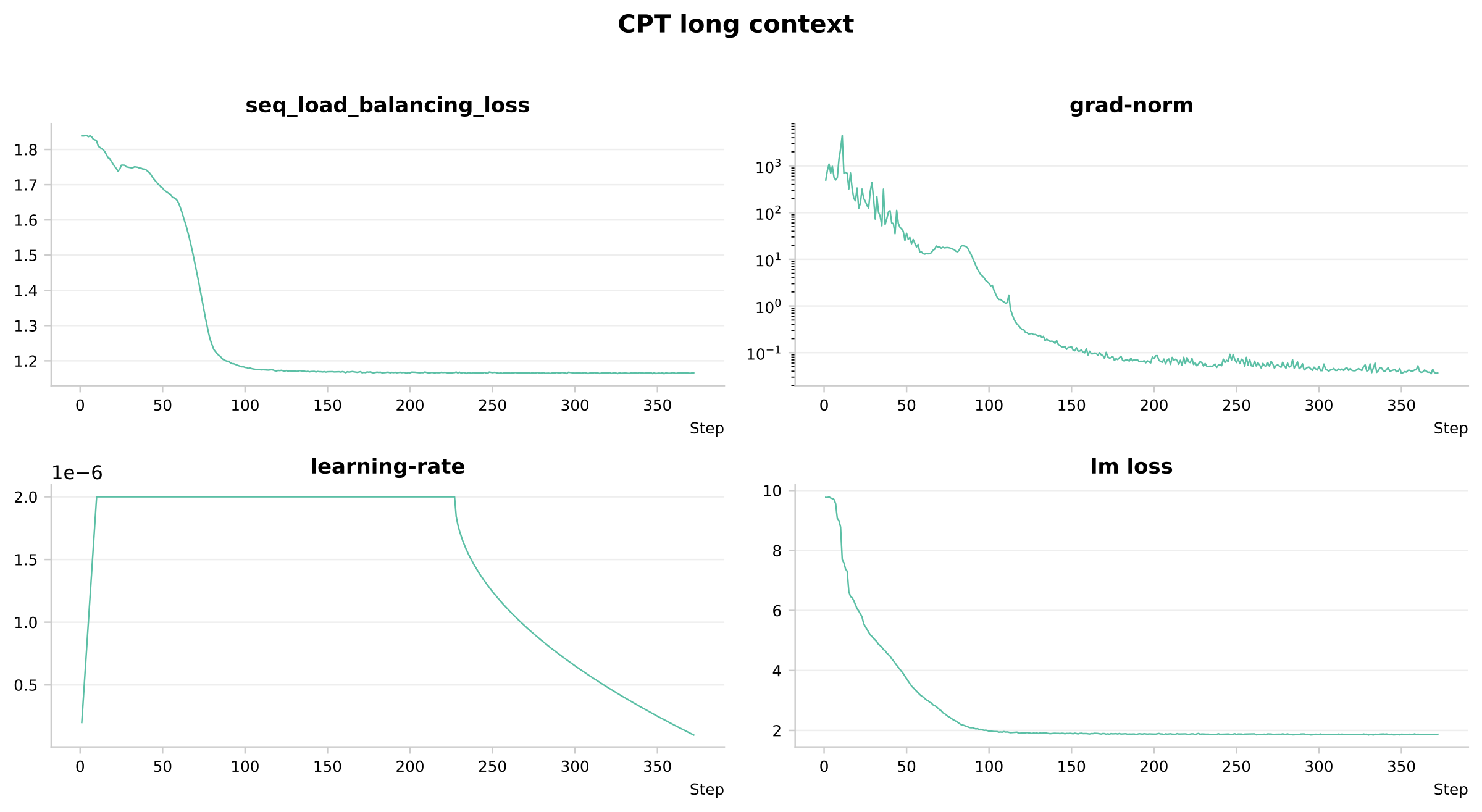}
  \caption{Weights \& Biases traces for long-context training at sequence length 131{,}072. Gradient norm and language-model loss fall early in training and then stabilize. Sequence load-balancing loss settles near 1.17 during the recorded run.}
  \label{fig:cptlongwandb}
\end{figure}

\section{Broader impacts}
\label{app:impacts}

The expanded tokenizer represents native-script text in eight Indian languages with fewer tokens. This allows more text to fit within a fixed token budget, although we have not measured end-to-end latency or serving costs. The public model weights allow other groups to build on the adaptation without repeating the full training process.

The adaptation reduced accuracy on several English benchmarks (Section~\ref{sec:cost}). Applications that require strong English performance should consider these losses when selecting a checkpoint. Romanized text remained more fragmented than native-script text, so users typing Indic languages in Latin script may not receive the same tokenization benefits.

The model could also be misused to generate misinformation in the supported languages. This is a concern where detection and moderation tools are limited. Odia, Punjabi, Urdu, and Nepali received a small share of the pretraining corpus but no dedicated vocabulary expansion. Benefits for languages outside the eight expanded here remain untested.

\section{Reproducibility}
\label{app:repro}

Repeating the full adaptation requires substantial compute. Our CPT runs used 256 H200 GPUs. The reusable components include the tokenizer and its construction procedure, initialization from existing token rows, checks that inherited rows remain unchanged, data-mixture specifications, and evaluation settings. Tokenizer and row-freezing tests do not require a full CPT run.

\clearpage
\endgroup

\end{document}